\pdfoutput=1

\documentclass[11pt]{article}

\usepackage[]{acl}

\usepackage{times}
\usepackage{latexsym}
\usepackage[T1]{fontenc}
\usepackage[utf8]{inputenc}
\usepackage{microtype}

\usepackage{xspace}
\usepackage{graphicx}
\usepackage{multirow}
\usepackage{tcolorbox}
\usepackage{booktabs}
\usepackage{amsmath}
\usepackage{amssymb}
\usepackage[normalem]{ulem}
\usepackage{xcolor}
\usepackage{colortbl}
\usepackage{utfsym}
\usepackage{pifont}
\usepackage{wasysym}
\usepackage{makecell}
\usepackage{fontawesome}
\usepackage{threeparttable}
\usepackage{enumitem}

\usepackage{longtable}

\def\eg{\textit{e.g.,} }
\def\ie{\textit{i.e.,} }
\def\bench{EvoCodeBench\xspace}

\useunder{\uline}{\ul}{}

\title{\bench: An Evolving Code Generation Benchmark Aligned with Real-World Code Repositories}

\author{
    Jia Li $\male^{1}$, Ge Li$^{1}$, Xuanming Zhang$^{1}$, Yihong Dong$^{1}$, Zhi Jin$^{1}$ \\
    $^1$School of Computer Science, Peking University  \\
    \texttt{lijia@stu.pku.edu.cn, lige@pku.edu.cn} \\
}

\begin{document}

\maketitle

\begin{abstract}
How to evaluate Large Language Models (LLMs) in code generation is an open question. 
Existing benchmarks demonstrate poor alignment with real-world code repositories and are insufficient to evaluate the coding abilities of LLMs.
This paper proposes a new benchmark - \textbf{\bench} to address the preceding problems, which has three primary advances.
\ding{182} \bench aligns with real-world repositories in multiple dimensions, \eg code distributions and dependency distributions.
\ding{183} \bench offers comprehensive annotations (\eg requirements, reference code, and reference dependencies), and robust evaluation metrics (\eg Pass@$k$ and Recall@$k$).
\ding{184} \bench is an evolving benchmark to avoid data leakage. We build an automatic pipeline to update \bench from the latest repositories.
We release the first version - \bench-2403, containing 275 samples from 25 real-world repositories.
Based on \bench, we propose \textbf{repository-level code generation} and evaluate 10 popular LLMs (\eg gpt-4, gpt-3.5, DeepSeek Coder, StarCoder 2, CodeLLaMa, Gemma, and Qwen 1.5). 
Our experiments reveal the coding abilities of these LLMs in real-world repositories. 
\textbf{For example, the highest Pass@1 of gpt-4 only is 20.73\% in our experiments.} 
We also analyze failed cases and summarize the shortcomings of existing LLMs in \bench.
We release \bench, all prompts, and LLMs' completions for further community analysis\footnote{\url{https://github.com/seketeam/EvoCodeBench}}.
\end{abstract}

\section{Introduction}
\label{sec:Introduction}

Code generation with Large Language Models (LLMs) has attracted lots of researchers' attention \cite{DeepSeek_Coder, CodeLLaMa, StarCoder-2}, and some commercial products have been produced, \eg GitHub Copilot \cite{Copilot}. In practice, human developers typically write the code for code repositories. Thus, evaluating the coding abilities of LLMs in real-world repositories is necessary. We analyze over 1 million functions from 500 real-world repositories (see Section \ref{sec:benchmark_collection}) and think a good benchmark should satisfy the following features.

\begin{figure}[t]
\centering
\includegraphics[width=0.9\linewidth]{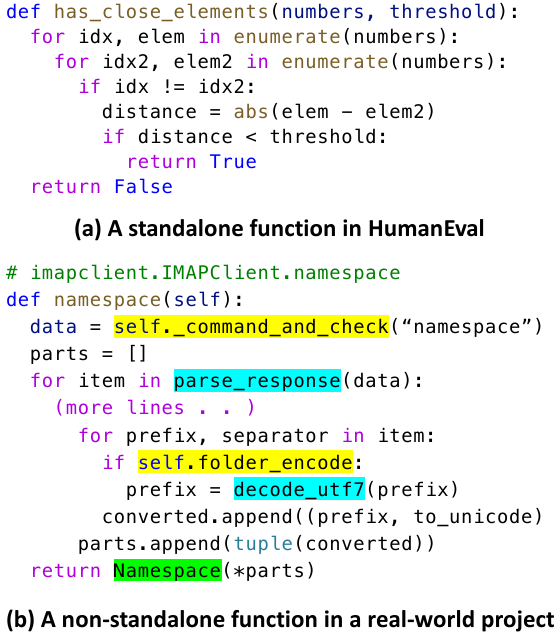}
\caption{Examples of standalone and non-standalone functions. Dependencies are highlighted, \ie yellow: intra-class dependencies, green: intra-file dependencies, and blue: cross-file dependencies.}
\label{fig:Introduction_Example}
\end{figure}

\begin{table*}[t]
\caption{The comparison between existing benchmarks and \bench.}
\label{tab:feature_comparison}
\resizebox{\linewidth}{!}{
\begin{tabular}{lccccc}
\toprule
Benchmark & Real Repo. & Real Code Distribution & Comprehensive Annota. & Robust Metric & Avoiding Data Leak. \\ \midrule
CoNaLA \cite{CoNaLA} & \faTimes & \faTimes & \faTimes & \faTimes & \faTimes \\
Concode \cite{Concode} & \faCheck & \faTimes & \faTimes & \faTimes & \faTimes \\
HumanEval \cite{Codex} & \faTimes & \faTimes & \faTimes & \faTimes & \faTimes \\
MBPP \cite{MBPP} & \faTimes & \faTimes & \faTimes & \faTimes & \faTimes \\
APPS \cite{APPS} & \faTimes & \faTimes & \faTimes & \faTimes & \faTimes \\
PandasEval \cite{CERT} & \faTimes & \faTimes & \faTimes & \faTimes & \faTimes \\
NumpyEval \cite{CERT} & \faTimes & \faTimes & \faTimes & \faTimes & \faTimes \\
AixBench \cite{SkCoder} & \faCheck & \faTimes & \faTimes & \faTimes & \faTimes \\
ClassEval \cite{ClassEval} & \faTimes & \faTimes & \faTimes & \faTimes & \faTimes \\
CoderEval \cite{CoderEval} & \faCheck & \faTimes & \faTimes & \faCheck & \faTimes \\ \midrule
\bench (Ours) & \faCheck & \faCheck & \faCheck & \faCheck & \faCheck \\
\bottomrule
\end{tabular}}
\end{table*}

\begin{itemize}[leftmargin=*]
    \item \textbf{Real-world Repository.} 
    The benchmark should be collected from real-world code repositories \cite{CoderEval}.
    \item \textbf{Real Code Distribution.} Real-world repositories comprise two types of code, \ie standalone and non-standalone code.
    As shown in Figure \ref{fig:Introduction_Example}, a standalone function solely uses built-in or public libraries, while a non-standalone one contains context-aware \textit{dependencies} (\ie invocations of code elements defined in current repositories). The benchmark should cover both types of code and ensure their ratios are realistic. The number of dependencies should also be consistent with real-world repositories.
    \item \textbf{Comprehensive Annotations.} The benchmark can offer comprehensive annotations, including natural language requirements, original repositories, and ground truths (code and dependencies).
    \item \textbf{Robust Evaluation Metrics.} The benchmark should contain test cases to evaluate models' predictions and report Pass@$k$. Metrics are also required to assess the abilities of LLMs to generate dependencies.
    \item  \textbf{Avoidng Data Leaking.} With more LLMs emerging, the benchmark should avoid potential data leakage \cite{competition_problems}.
\end{itemize}
However, as shown in Table \ref{tab:feature_comparison}, none of the existing benchmarks satisfies all aforementioned features.
The problem hinders the evaluation and development of LLMs in the real development process.

\textbf{To address the above problem, we propose a new code generation benchmark named \bench, which aligns with real-world code repositories.} As shown in Table \ref{tab:feature_comparison}, \bench satisfies the above features. \ding{182} \bench is collected from high-quality open-source repositories in the real world. \ding{183} \bench is constructed through a rigorous pipeline and aligns with real-world repositories. Specifically, the distributions of code and dependencies in \bench are consistent with the ones in 500 real-world repositories. Detailed statistics are in Section \ref{sec:bench:2403}. \ding{184} \bench offers comprehensive annotations, \eg detailed requirements, original repositories, reference code, and reference dependencies. \ding{185} \bench leverages test cases to check models' predictions and report Pass@$k$. It also proposes Recall@$k$ to evaluate the dependencies in predictions. \ding{186} \bench is an evolving benchmark and will be dynamically updated every period (\eg 6 months) to avoid data leakage. In this paper, we release the first version - \bench-2403, which consists of 275 samples from 25 real-world repositories.
Details of the collection process can be found in Section \ref{sec:benchmark_collection}.

Based on \bench, we propose \textbf{repository-level code generation}, which simulates the developers' coding process in a working repository. The task asks models to write the code based on requirements and a complete repository.

\begin{figure*}[t]
\centering
\includegraphics[width=0.85\linewidth]{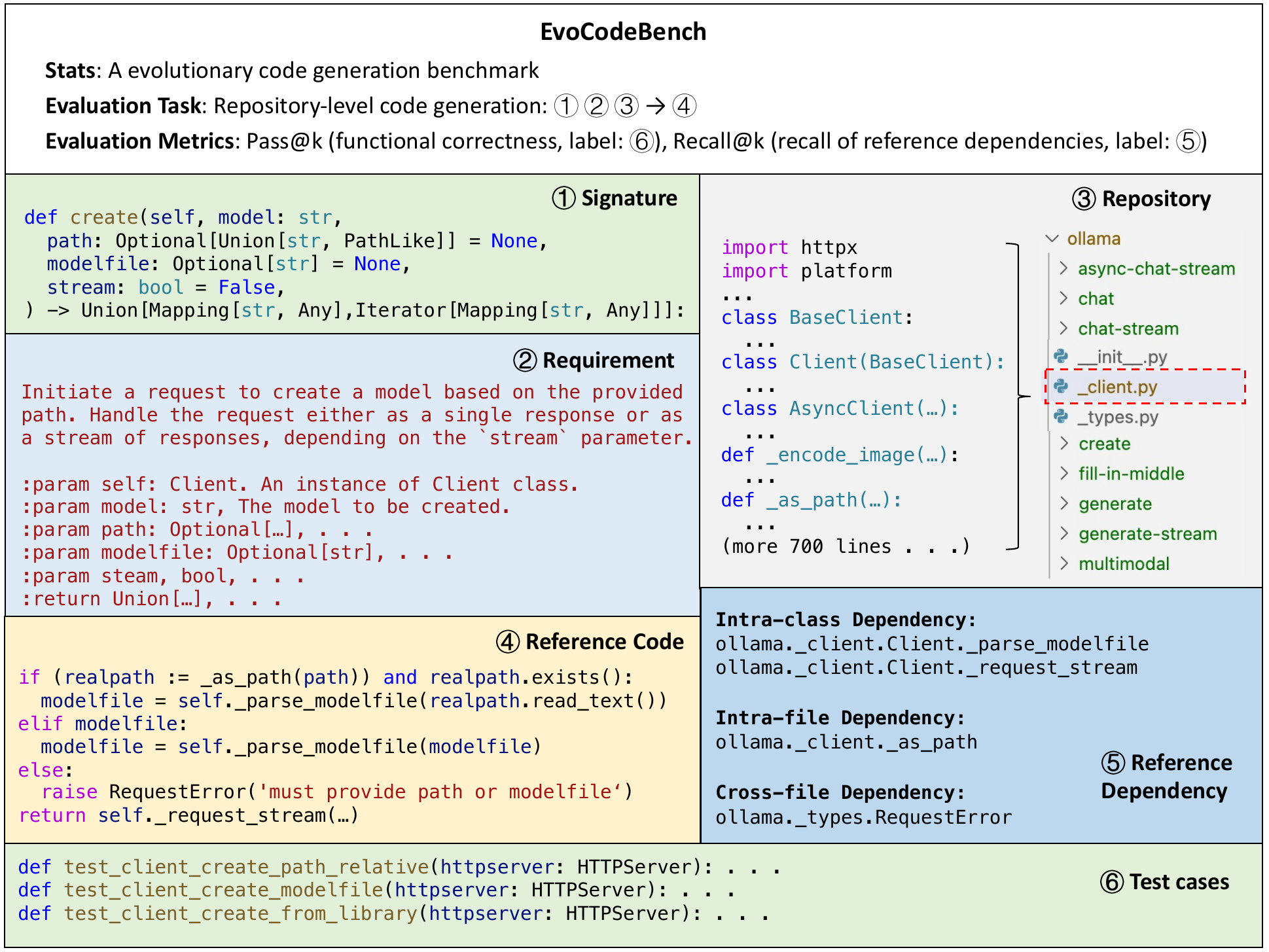}
\caption{An overview of \bench. Each sample consists of six components.}
\label{fig:Benchmark_example}
\end{figure*}

We evaluate 10 popular LLMs (\ie gpt-4 \cite{gpt-4}, gpt-3.5 \cite{gpt-3.5}, DeepSeek Coder \cite{DeepSeek_Coder}, StarCoder 2 \cite{StarCoder-2}, CodeLLaMa \cite{CodeLLaMa}, Gemma \cite{gemma}, and Qwen 1.5 \cite{Qwen}). These LLMs exhibit low performance on \bench, especially compared to their performance on previous benchmarks. \textbf{For example, gpt-4-turbo-1106 achieves a Pass@1 score of 80\% on HumanEval, while its highest Pass@1 on \bench is only 20.73\%.} Our results reveal the coding abilities of these LLMs in real-world repositories. We further analyze failed cases and summarize the shortcomings of existing LLMs
in \bench.

In summary, our contributions are as follows:
\begin{itemize}[leftmargin=*]
    \item We summarize five features (see Table \ref{tab:feature_comparison}) that a code generation benchmark for real-world repositories should satisfy.
    \item We propose a new code generation benchmark - \bench, satisfying the above features. We released the first version and will continually update it.
    \item We propose repository-level code generation, which provides a challenging and realistic evaluation scenario.
    \item We evaluate 10 popular LLMs on \bench, analyzing their strengths and shortcomings in repository-level code generation.
\end{itemize}

We hope \bench can align with the actual experiences of developers during the practical development process. By \bench, practitioners can pick up superior LLMs and facilitate the application of code generation techniques in real-world repositories.

\section{\bench}
\label{sec:bench}

In this section, we first show an overview of \bench and then describe its tasks and metrics. Finally, we present the first version - \bench-2403 and its statistics.

\subsection{Overview}
\label{sec:bench:overview}

Figure \ref{fig:Benchmark_example} shows a sample in \bench. Each sample consists of six components. 

\textbf{\ding{182} Function Signature:} The signature of the target code. 
\textbf{\ding{183} Requirement:} An English description detailing the functionality of the target code. 
\textbf{\ding{184} Repository:} The current repository contains hundreds of code files.
\textbf{\ding{185} Reference Code:} A developer-written implementation of the target code. This code may invoke dependencies defined in the current repository.
\textbf{\ding{186} Reference Dependency:} The dependencies invoked in the reference code include intra-class, intra-file, and cross-file dependencies.
\textbf{\ding{187} Test Cases:} Test cases are used to check the functional correctness of the code.

\subsection{Task Definition}
\label{sec:bench:tasks}

Traditional benchmarks fall into a simple requirement-to-code task. In contrast, \bench proposes a more realistic task - \textbf{repository-level code generation}. This task simulates the developers' coding process in a working repository. Given a requirement and a repository, LLMs are tasked to generate the code for the repository. 

\begin{table*}[t]
\caption{The comparison between existing code generation benchmarks and \bench-2403. \texttt{SA} and \texttt{Depend} are the abbreviations of ``standalone'' and ``dependency'', respectively.}
\label{tab:benchmark_comparison}
\resizebox{\linewidth}{!}{
\begin{tabular}{l|cccc|cc|c|c}
\toprule
\multirow{2}{*}{Benchmark} & \multicolumn{4}{c}{Code Distribution} & \multicolumn{2}{|c|}{Dependency Distribution} & \multirow{2}{*}{Annotation} & \multirow{2}{*}{Release Data} \\ 
 & \#Repo. & \#Sample & SA & Non-SA & \#Type & \#Avg. &  &  \\ \midrule
CoNaLa & -- & 500 & 100\% & 0\% & 0 & 0 & NL, Code & \texttt{2018-5} \\
HumanEval & -- & 164 & 100\% & 0\% & 0 & 0 & NL, Code & \texttt{2021-7} \\
MBPP & -- & 974 & 100\% & 0\% & 0 & 0 & NL, Code & \texttt{2021-8} \\
APPS & -- & 5,000 & 100\% & 0\% & 0 & 0 & NL, Code & \texttt{2021-11} \\
PandasEval & -- & 101 & 100\% & 0\% & 0 & 0 & NL, Code & \texttt{2022-6} \\
NumpyEval & -- & 101 & 100\% & 0\% & 0 & 0 & NL, Code & \texttt{2022-6} \\
AixBench & N/A & 175 & 100\% & 0\% & 0 & 0 & NL, Code & \texttt{2023-2} \\
ClassEval & -- & 100 & 100\% & 0\% & 0 & 0 & NL, Code, Depend. Name & \texttt{2023-8} \\
\midrule
Concode & N/A & 2,000 & 20\% & 80\% & 1 & 1.23 & NL, Code & \texttt{2018-8} \\
CoderEval & 43 & 230 & 36\% & 64\% & 3 & 1.73 & NL, Code, Depend. Name & \texttt{2023-2} \\
\rowcolor[rgb]{ .741,  .843,  .933}
EvoCodeBench-2403 & 25 & 275 & 27\% & 73\% & 3 & 3.46 &\begin{tabular}[c]{@{}c@{}}NL, Code, Depend. Path,\\ Repository\end{tabular} & \texttt{2024-3} \\ \midrule
500 Real Repositories & 500 & 1M+ & 27\% & 73\% & 3 & 3.22 & -- & -- \\
\bottomrule
\end{tabular}}
\end{table*}

\subsection{Evaluation Metrics}
\label{sec:bench:metric}

\textbf{Pass@$k$ (Functional Correctness).} Following previous studies \cite{Codex,MBPP,CoderEval}, we assess the functional correctness of programs by executing test cases and compute the unbiased Pass@$k$.
Specifically, we generate $n \geq k$ programs per requirement, count the number of correct programs $c \leq n$ that pass test cases, and calculate the Pass@$k$:
\begin{equation}
\text{Pass}@k:=\underset{\text { Requirements }}{\mathbb{E}}\left[1-\frac{\left(\begin{array}{c}
n-c \\
k
\end{array}\right)}{\left(\begin{array}{l}
n \\
k
\end{array}\right)}\right]
\end{equation}

\textbf{Recall@$k$ (Recall of Reference Dependency).} 
Besides the functional correctness, we expect LLMs to invoke relevant dependencies defined in contexts. Hence, we propose Recall@$k$, which gauges the recall of reference dependencies in generated programs.  

Specifically, LLMs generate $k$ programs per requirement. For the $i$-th program, we employ a parser\footnote{We develop the parser based on an open-source static analysis tool - Pyan \cite{Pyan}.} to extract its dependencies as $\mathbb{P}_i$. Subsequently, we compare $\mathbb{P}_i$ with reference dependencies $\mathbb{R}$ and compute the Recall@$k$:
\begin{equation}
\text{Recall}@k:= \underset{\text{Requirements}}{\mathbb{E}} \left[ \max_{i \in [1, k]} \frac{|\mathbb{R} \cap \mathbb{P}_i|}{|\mathbb{R}|}\right]
\end{equation}
where $|\cdot|$ means the number of elements of a set.

\subsection{\bench-2403}
\label{sec:bench:2403}

This paper releases the first version of \bench named \bench-2403. The statistics of \bench-2403 are shown in Table \ref{tab:benchmark_comparison}. We discuss its features as follows.

\ding{182} \textbf{Alignment with real-world code repositories.} 
\bench-2403 consists of 275 samples collected from 25 real-world repositories.
As shown in Table \ref{tab:benchmark_comparison}, the code distribution of \bench-2403 is consistent with that of 500 real-world repositories. The average number of dependencies per program in \bench-2403 is also close to that of 500 real-world repositories.

\ding{183} \textbf{Comprehensive Annotations.} We provide requirements, reference code, reference dependency and the complete repository for each sample.
Previous works (\ie CoderEval, ClassEval) only provide dependencies' names (\eg \texttt{close}).
Because many functions have the same name in practice, it is hard to identify whether generated dependencies are correct by relying on names. \bench annotates dependencies with paths (\eg \texttt{A.py::ClassB::close}), addressing ambiguity and biases. These annotations offer a broad arena to explore repository-level code generation and evaluation.

\ding{184} \textbf{Latest repositories to avoid data leakage.} Considering the latest LLM's \cite{StarCoder-2} training data is up to \texttt{2023-9}, \bench-2403 is collected from real-world repositories that were created from \texttt{2023-10} to \texttt{2024-2}. In future versions, we will continually update \bench using the latest repositories.

\section{Benchmark Collection Pipeline}
\label{sec:benchmark_collection}

We build an automatic pipeline for collecting \bench from the latest repositories. The pipeline consists of four stages as follows.

\textbf{\ding{182} Repository Selection.} We crawl high-quality repositories from GitHub satisfying the following criteria: open-source Python projects; created in recent months; non-fork and non-malicious projects; more than 50 stars; and having explicit unit tests. 

\textbf{\ding{183} Function Parse.} We extract functions from repositories and exclude trivial functions (\eg empty or initialization functions). We extract each function’s signature and function body (\ie reference code). We developed a static analysis-based parser to extract reference dependencies with the reference code.

\textbf{\ding{184} Tests Construction.} For each function, we extract test cases invoking it from current repositories. We use \texttt{pip}\footnote{\url{https://pypi.org/project/pip/}} to automatically install required packages for each repository and leverage \texttt{Pytest}\footnote{\url{https://docs.pytest.org/en/8.0.x/}} to run test cases. Functions without executable test cases are excluded. 

\textbf{\ding{185} Deduplication.} To improve the diversity of \bench, we perform repository-level deduplication based on the Jaccard similarities in code files and imports. The former removes duplicate repositories in text surfaces, and the latter removes repositories in domains that are too similar.

\textbf{\ding{186} Requirement Annotation.} Manually writing requirements is time-consuming and laborious. Inspired by the powerful abilities of LLMs in code comment generation \cite{LLM4CS}, we leverage LLMs to generate natural language requirements. Specifically, we manually craft a few-shot prompt, which teaches LLMs to write requirements in a specific format (\ie functional descriptions and input-output parameters). Appendix \ref{sec:appendix:collection:gen_requirement} the details of prompts.

\textbf{\ding{187} Benchmark Construction.} Finally, we select samples from the outputs of step \ding{186} to construct \bench. We strive to make \bench satisfy the following goals: consistent with the code distribution observed in 500 real-world repositories; close to the average number of dependencies in 500 real-world repositories; including as many samples as possible. 

\section{Experiments}
\label{sec:experiments}

\begin{table}[t]
\caption{Studied LLMs in this paper. Context L.: Context Window.}
\label{tab:Base_LLMs}
\resizebox{\linewidth}{!}{
\begin{tabular}{llcc}
\toprule
Type & Name & Version & Context W. \\ \midrule
\multirow{2}{*}{Closed-source} 
 & gpt-4 & gpt-4-turbo-1106 & 128,000 \\ 
 & gpt-3.5 & gpt-3.5-turbo-1106 & 16,385 \\ \midrule
\multirow{8}{*}{Open-source} 
 & StarCoder 2 & 15B & 16,384 \\
 & StarCoder 2 & 7B & 16,384 \\
 & DeepSeek Coder & 33B & 16,384 \\
 & DeepSeek Coder & 6.7B & 16,384 \\
 & CodeLLaMa & 13B & 16,384 \\
 & CodeLLaMa & 7B & 16,384 \\
 & Gemma & 7B & 8,192 \\
 & Qwen 1.5 & 7B & 32,768 \\
 \bottomrule
\end{tabular}}
\end{table}

\subsection{Studied LLMs}
\label{sec:experiments:base_llms}

Table \ref{tab:Base_LLMs} shows 10 studied LLMs in our experiments. They cover closed-source LLMs (\ie gpt-4 \cite{gpt-4}, gpt-3.5 \cite{gpt-3.5}) and open-source LLMs (\ie DeepSeek Coder \cite{DeepSeek_Coder}, StarCoder 2 \cite{StarCoder-2}, CodeLLaMa \cite{CodeLLaMa}, Gemma \cite{gemma}, and Qwen 1.5 \cite{Qwen}). We use official interfaces or implementations to reproduce these LLMs. The details of LLMs can be found in Appendix \ref{sec:appendix:experiment:llm}. 

\begin{table*}[t]
\caption{Pass@$k$ and Recall@$k$ of LLMs on \bench-2403. Bold and underlined data indicate top-1 and top-2 results, respectively.}
\label{tab:CGen}
\resizebox{\linewidth}{!}{
\begin{tabular}{lc|cccc|cccc}
\toprule
LLMs & Size & Pass@1 & Pass@3 & Pass@5 & Pass@10 & Recall@1 & Recall@3 & Recall@5 & Recall@10 \\
\midrule
\rowcolor[rgb]{ .741,  .843,  .933}
\multicolumn{10}{c}{Local File (Infilling)} \\
\midrule
gpt-4 & N/A & \textbf{20.73} & \textbf{23.03} & \textbf{24.11} & \textbf{25.34} & 68.24 & 70.63 & 72.05 & 73.52 \\
gpt-3.5 & N/A & 17.82 & 21.78 & 23.06 & 24.46 & 61.94 & 68.13 & 69.69 & 70.85 \\
DeepSeek Coder & 33B & {\ul 19.64} & {\ul 22.78} & {\ul 24.29} & {\ul 26.01} & \textbf{71.46} & \textbf{79.93} & \textbf{82.11} & \textbf{86.25} \\
DeepSeek Coder & 6.7B & 17.82 & 21.02 & 22.40 & 23.97 & {\ul 69.58} & {\ul 74.04} & {\ul 78.00} & {\ul 83.22} \\
StarCoder 2 & 15B & 14.91 & 17.54 & 18.63 & 19.86 & 50.90 & 53.29 & 55.89 & 61.76 \\
StarCoder 2 & 7B & 15.27 & 17.29 & 18.63 & 20.09 & 56.35 & 60.59 & 63.74 & 74.20 \\
\midrule
\rowcolor[rgb]{ .741,  .843,  .933}
\multicolumn{10}{c}{Local File (Completion)} \\
\midrule
gpt-4 & N/A & \textbf{17.45} & \textbf{19.65} & \textbf{20.80} & \textbf{22.41} & 63.49 & 68.67 & 70.00 & 72.07 \\
gpt-3.5 & N/A & {\ul 15.64} & 17.29 & 18.21 & 19.36 & 61.44 & 66.25 & 66.82 & 69.89 \\
DeepSeek Coder & 33B & 14.18 & {\ul 17.57} & 18.66 & 19.95 & {\ul 66.90} & \textbf{72.83} & 74.40 & 80.02 \\
DeepSeek Coder & 6.7B & 13.45 & 17.10 & {\ul 18.81} & {\ul 21.07} & 65.76 & {\ul 72.32} & {\ul 75.61} & 78.45 \\
StarCoder 2 & 15B & 13.45 & 15.44 & 17.84 & 19.59 & \textbf{68.55} & 71.37 & 74.76 & 77.70 \\
StarCoder 2 & 7B & 13.82 & 15.15 & 16.18 & 17.65 & 62.93 & 69.85 & 73.54 & 78.40 \\
CodeLLaMa & 13B & 12.73 & 15.78 & 16.86 & 18.19 & 63.34 & 71.26 & \textbf{76.43} & {\ul 80.11} \\
CodeLLaMa & 7B & 12.73 & 15.33 & 16.00 & 16.93 & 63.33 & 69.79 & 71.91 & 76.50 \\
Gemma & 7B & 10.55 & 13.25 & 14.31 & 15.48 & 58.02 & 70.57 & 74.44 & \textbf{80.76} \\
Qwen 1.5 & 7B & 5.45 & 7.04 & 7.91 & 9.07 & 39.21 & 44.02 & 50.17 & 58.42 \\
\midrule
\rowcolor[rgb]{ .741,  .843,  .933}
\multicolumn{10}{c}{Without Contexts} \\ \midrule
gpt-4 & N/A & \textbf{7.27} & \textbf{10.05} & \textbf{10.70} & {\ul 11.49} & 21.58 & 23.93 & 25.69 & 26.23 \\
gpt-3.5 & N/A & 6.55 & 7.85 & 8.28 & 8.73 & 21.66 & 24.31 & 24.77 & 25.40 \\
DeepSeek Coder & 33B & {\ul 6.91} & {\ul 8.92} & 9.79 & 11.03 & \textbf{27.67} & \textbf{32.73} & {\ul 34.92} & {\ul 37.76} \\
DeepSeek Coder & 6.7B & 5.82 & 8.56 & 9.67 & 11.26 & 25.89 & 32.06 & \textbf{35.59} & \textbf{38.33} \\
StarCoder 2 & 15B & 6.18 & 8.77 & {\ul 9.95} & \textbf{11.53} & 24.03 & 29.86 & 33.62 & 36.91 \\
StarCoder 2 & 7B & 5.82 & 6.72 & 7.43 & 8.62 & {\ul 27.39} & {\ul 32.60} & 34.88 & 36.81 \\
CodeLLaMa & 13B & 5.45 & 7.38 & 8.37 & 9.95 & 25.52 & 31.28 & 33.66 & 36.36 \\
CodeLLaMa & 7B & 5.45 & 6.94 & 7.75 & 9.03 & 26.97 & 31.17 & 34.08 & 36.82 \\
Gemma & 7B & 6.18 & 6.86 & 7.64 & 8.66 & 21.84 & 29.98 & 33.61 & 35.23 \\
Qwen 1.5 & 7B & 4.00 & 4.72 & 5.38 & 6.18 & 16.33 & 13.56 & 16.34 & 21.06 \\
\bottomrule
\end{tabular}}
\end{table*}

\subsection{Experimental Setting}
\label{sec:experiments:setting}

Repository-level code generation takes a requirement and a repository as inputs. Typically, a repository consists of hundreds of code files and is very long. For example, the average length of 500 real-world repositories is 1.1 million tokens, surpassing the context windows of existing LLMs (\eg gpt-4: 128k tokens). Inspired by related work \cite{Repo_prompt}, we try to extract parts of code contexts from the repository as inputs and design the following experimental settings.

\textbf{\ding{182} Without context.} In this setting, we ignore contexts and directly generate the code based on requirements and signatures.

\textbf{\ding{183} Local File (Completion).} The local file denotes the code file where the reference code is in. This setting simulates the scenario where developers continue to write code at the end of a file. Thus, we consider code snippets above the reference code in the local file as contexts. Then, LLMs generate code in an autoregressive manner based on requirements, signatures, and contexts.

\textbf{\ding{184} Local File (Infilling).} Different from the Local File (Completion) setting, this setting simulates the scenario where developers infill code in the middle of a file. Thus, we use the code snippets above and below the reference code in the local file as contexts. We evaluate LLMs that support code infilling and construct input sequences using official formats.

The prompt templates used in the above settings are shown in Appendix \ref{sec:appendix:experiment:prompt}. We note that there are other approaches to extracting relevant contexts. We consider this to be beyond the scope of this paper and leave the exploration to future work.

\subsection{Evaluation}
\label{sec:experiments:evaluation}

We use Pass@$k$ and Recall@$k$ (see Section \ref{sec:bench:metric}) to assess generated programs. In this paper, $k \in [1, 3, 5, 10]$. When $k=1$, we use the greedy search and generate a single program per requirement. When $k>1$, we use the nucleus sampling with a temperature 0.4 and sample 20 programs per requirement. We set the top-$p$ to 0.95 and the max generation length to 500.

Because \bench is an evolving benchmark, this paper evaluates LLMs upon \bench-2403. Note that the Pass@$k$ and Recall@$k$ between different versions of \bench are not comparable.

\subsection{Main Results}
\label{sec:experiments:main_results}

The Pass@$k$ and Recall@$k$ in three experimental settings are shown in Table \ref{tab:CGen}.

\noindent \textbf{Without Context.} gpt-4 achieves the highest Pass@$k$ among all LLMs. However, compared to previous benchmarks, these LLMs' performance in \bench-2403 drops dramatically. For example, gpt-4 achieves a Pass@1 score of 88.4 on HumanEval, while it scores 7.27 on Pass@1 upon \bench-2403. On the one hand, the decreases validate our motivation that existing benchmarks can not comprehensively assess the coding abilities of LLMs in practical projects. On the other hand, the results emphasize the importance of contexts in repositories. Without the necessary context, LLMs lack the domain knowledge of current repositories and generate the wrong programs.

\begin{figure}[t]
\centering
\includegraphics[width=\linewidth]{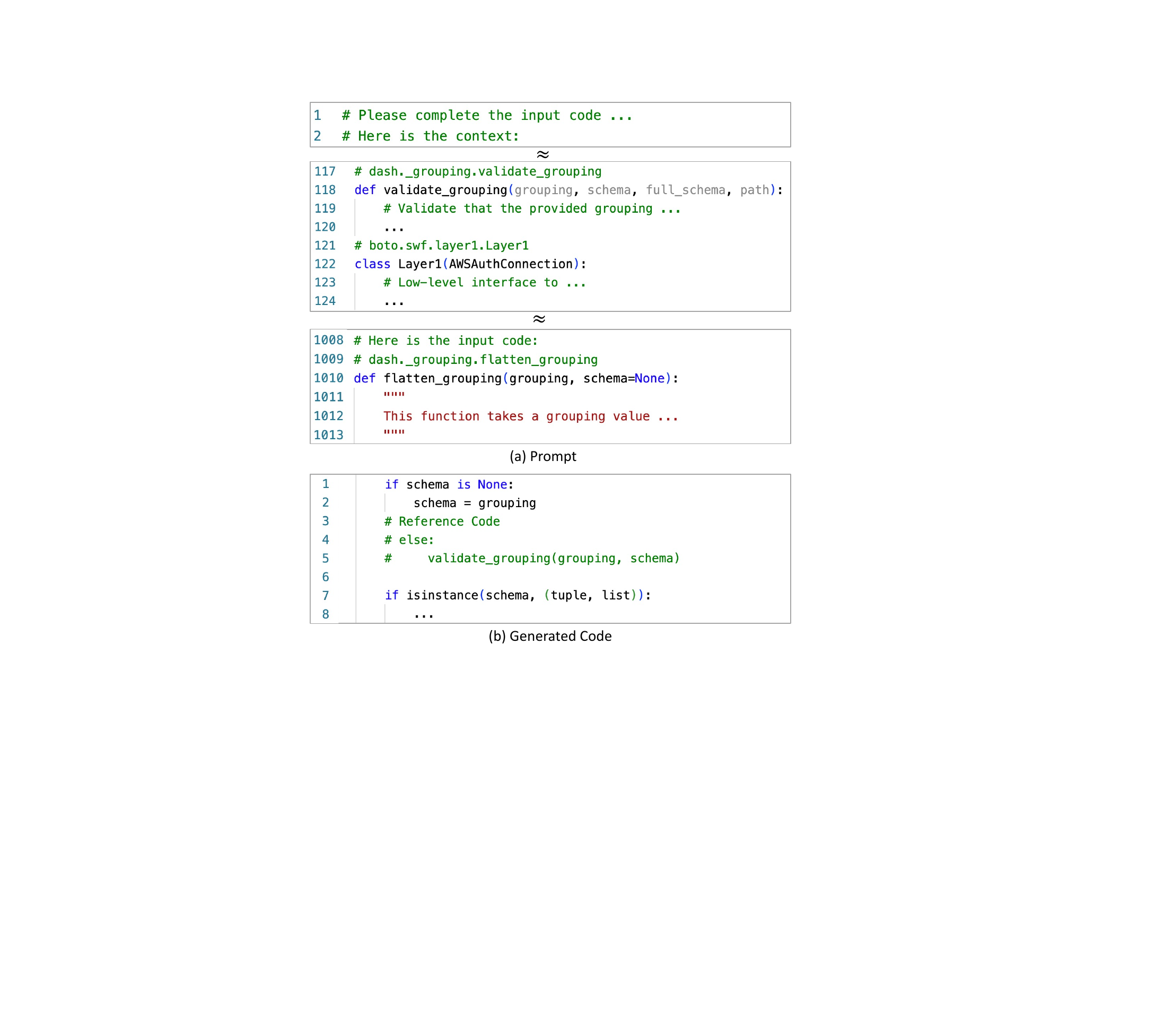}
\vspace{-0.2cm}
\caption{A uniquely successful case in Local File (Completion) setting.}
\label{fig:success_case}
\end{figure}

\noindent \textbf{Local File (Completion) and (Infilling).} After introducing the contexts within local files, the Pass@$k$ and Recall@$k$ of all LLMs obviously increase. For example, the Pass@1 of gpt-4 is improved by 104\% and 152\% in two settings, respectively. We attribute the improvements to the domain knowledge contained in contexts. Figure \ref{fig:success_case} shows a uniquely successful case in the Local File (Completion) setting. The key to writing this function is to know cache directories. Without context, gpt-4 fabricated a non-existent field as cache directories, generating the incorrect code. In fact, two functions for returning the cache directories are available in the local file. After introducing the local file, gpt-4 successfully gets cache directories by invoking these functions and generates the correct code.

\noindent \textbf{Error Analyses.} Although promising, the Pass@$k$ of existing LLMs is still low and far from practical applications. To determine LLMs' shortcomings, we manually analyze 50 error cases of gpt-4 in the Local File (Infilling) setting. We found that most of the cases (29 cases) failed due to implementation logic errors. 20 cases failed since the necessary contexts were missing, \eg APIs defined in other files. Besides, one case failed because of the vague requirement. It shows that existing LLMs' reasoning and coding abilities need to be improved. Meanwhile, how to utilize more contexts is necessary to explore.

We also obtain some interesting findings from Table \ref{tab:CGen}. 

\ding{182} \textbf{LLMs successfully generate some dependencies without context.} Theoretically, LLMs do not see the contexts and cannot generate dependencies. According to Table \ref{tab:CGen}, we are surprised to find that LLMs can generate some dependencies without context. We manually inspect successful cases and summarize two reasons. First, LLMs can reason about some easy dependencies from requirements, \eg initialization functions of returned objects. Second, LLMs can ``guess'' dependencies from their functionalities. In practice, dependencies' names are relevant to their functional descriptions, \eg \texttt{send\_request()} means send a request to the server. LLMs are trained with a large code corpus and can learn the naming conventions. Thus, LLMs may successfully guess some dependencies from their functionalities.

\ding{183} \textbf{More contexts benefit code generation.}
Based on Table \ref{tab:CGen}, we compare the performance of an LLM (\eg gpt-4) under different settings. Obviously, the more input contexts, the better the performance of the LLM. It inspires practitioners to extend the context windows of LLMs and input more contexts.

\ding{184} \textbf{The gpt family models have higher Pass@$k$ and lower Recall@$k$, while other models are the opposite.} We speculate the reason is that gpt family models are instruction-tuned models and focus on performing tasks based on given instructions. With limited contexts, gpt family models are conservative and tend to generate code independently. Other LLMs are standard language models trained with real code files containing dependencies. They are aggressive and generate dependencies that may exist. The comparisons show the importance of instruction tuning in practical applications.

\begin{figure}[t]
\centering
\includegraphics[width=\linewidth]{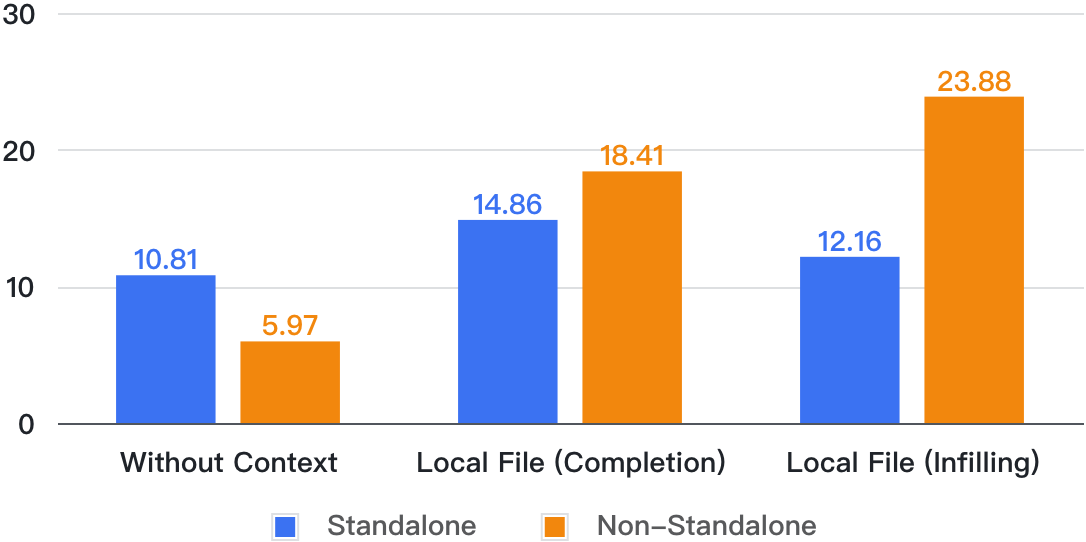}
\vspace{-0.2cm}
\caption{Pass@1 of gpt-4 on different program types.}
\label{fig:different_program}
\end{figure}

\subsection{Empirical Leassons}
\label{sec:experiments:lessons}

Based on the above experiments, we summarize the empirical lessons we learned as

\ding{182} \bench pose new challenges, \ie repository-level code generation. The performance of existing LLMs on \bench-2403 drops dramatically compared to their performance on previous benchmarks.

\ding{183} LLMs benefit from code contexts in current repositories. With limited context windows, the contexts from local files can improve gpt-4 by 152\% in Pass@1.

\ding{184} The main reasons why programs generated by LLMs fail are logic errors and incomplete contexts. How to enhance reasoning abilities and context windows of LLMs is important.

\section{Discussion}
\label{sec:discussion}

\begin{table}[t]
\caption{The comparison between auto-generated requirements and human-written requirements.}
\label{tab:requirement_evalution}
\resizebox{\linewidth}{!}{
\begin{tabular}{lccc}
\toprule
Annotator & Win / Tie / Lose & Cost (Time) & Cost (Money) \\
\midrule
gpt-4 & 5 / 41 / 4 & 12m30s & \$0.54 \\
Human & 4 / 41 / 5 & 4h10m & \$31.25 \\
\bottomrule
\end{tabular}}
\end{table}

\noindent \textbf{Evaluation of auto-generated requirements.} We leverage an LLM (gpt-4 in this paper) to generate natural language requirements for functions automatically. To assess the quality of auto-generated requirements, we randomly select 50 functions from \bench-2403 and compare requirements from gpt-4 and human developers. We hire two developers to write the requirements and two others to evaluate auto-generated and human-written requirements. The evaluation metrics include \textit{completeness} (whether the requirements cover the intent of code), \textit{clarity} (whether the requirements are clear and user-friendly). 
All developers are paid according to the relevant policies\footnote{\url{https://www.worker.gov/}} (\$7.5 per hour). 

The evaluation results are shown in Table \ref{tab:requirement_evalution}. The Cohen's Kappa coefficient between the two evaluators is 0.92.
On 41 functions, gpt-4 and developers tie. In the remaining functions, gpt-4 wins by 5 functions, and developers win by 4 functions. These results show that gpt-4 can produce high-quality requirements comparable to human-written requirements in most cases (92\% = 46/50). We also inspect the four functions lost by gpt-4 and find that some necessary details (\eg hyper-parameters) are missing in its requirements. In the future, we will explore new techniques to solve this problem, \eg controllable text generation \cite{control_gen}.
Besides the high-quality requirements, gpt-4 shows advantages in costs. As shown in Table \ref{tab:requirement_evalution}, got-4 costs less time and money to annotate requirements. Thus, it is a feasible and efficient approach for us to use gpt-4 to annotate requirements for \bench.

\noindent \textbf{Retrieval-Augmented Generation (RAG).} RAG is to enhance generative models with retrieved information and has achieved promising results in code generation \cite{SkCoder,AceCoder}. We try to apply RAG to repository-level code generation and consider the repository to be a retrieval corpus. Because most programs in repositories are not equipped with documentation, we retrieve top-$k$ (\ie $k=5$ in this paper) functions with similar names to the target function. Specifically, we split names into tokens based on underscore or camelcase formatting and then match the tokens of names. Finally, we use similar functions as contexts in prompts and further generate code. The experimental results are shown in Table \ref{tab:RAG}. The performance of both LLMs is improved after introducing similar functions. LLMs can know relevant algorithms and dependencies from similar functions, which benefit writing new programs. In the future, we will explore more advanced RAG techniques to improve repository-level code generation.

\begin{table}[t]
\centering
\caption{Pass@1 and Recall@1 with retrieval-augmented generation.}
\label{tab:RAG}
\resizebox{0.95\linewidth}{!}{
\begin{tabular}{llcc}
\toprule
LLMs & Setting & Pass@1 & Recall@1 \\
\midrule
\multirow{2}{*}{gpt-4} & Without Context & 8.31 & 21.08 \\
 & Similar Functions & 12.29 & 45.14 \\ \midrule
\multirow{2}{*}{gpt-3.5} & Without Context & 6.64 & 21.16 \\
 & Similar Functions & 11.62 & 41.93 \\
 \bottomrule
\end{tabular}}
\end{table}

\noindent \textbf{Results on different program types.} Figure \ref{fig:different_program} shows Pass@1 of gpt-4 on different program types (\ie standalone and non-standalone). The results are consistent with the above Table \ref{tab:CGen}. Code contexts significantly improve the Pass@1 on non-standalone functions. Meanwhile, the Pass@1 on standalone functions also slightly increases. We speculate that the domain knowledge (\eg private objects) within contexts helps LLMs understand requirements. However, the Pass@1 on both types of programs is still low. The coding abilities of existing LLMs in real-world repositories need to be further improved.

\begin{figure}[t]
\centering
\includegraphics[width=\linewidth]{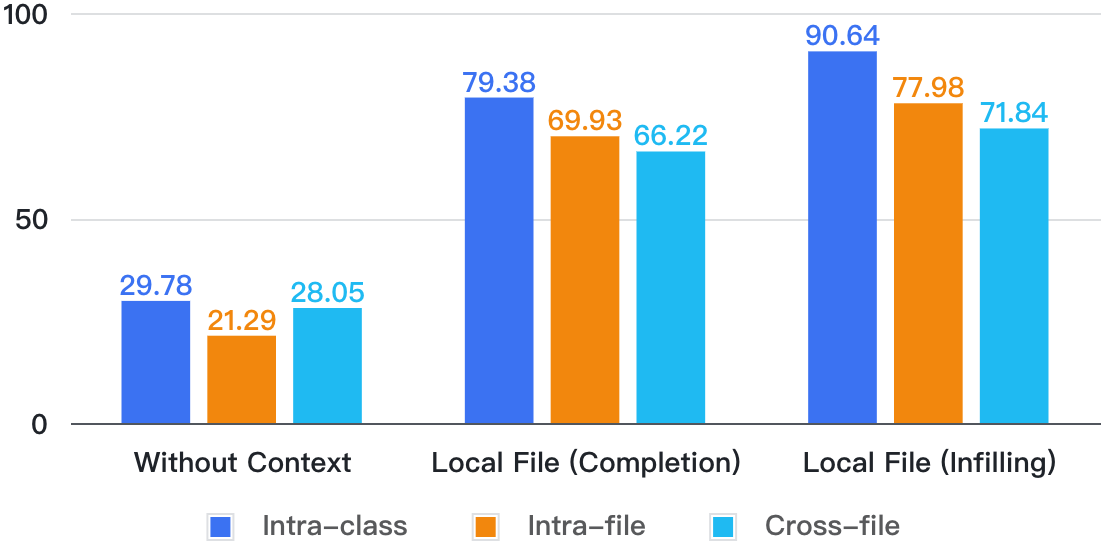}
\caption{Recall@1 of gpt-4 on different dependency types.}
\vspace{-0.2cm}
\label{fig:different_depend}
\end{figure}

\noindent \textbf{Results on different dependency types.} Figure \ref{fig:different_depend} shows the Recall@1 of gpt-4 on different dependency types (\ie intra-class, intra-file, and cross-file). The results yield two insights. \ding{182} Without context, LLMs can reason about some simple dependencies from requirements (\eg initialization functions of returned objects), but still exhibit low Recall@1 values across three dependency types. \ding{173} Local files contain code implementations of intra-class and intra-file dependencies and thus improve the Recall@1 on both types of dependencies.
It is surprising that Recall@1 on cross-file dependencies also increases. We inspect successful cases and find that LLMs \textit{copy} cross-file dependencies from relevant programs in local files. However, copied cross-dependencies may be inconsistent with the current code, \eg inconsistent arguments. It results in gpt-4 with higher Recall@1 but lower Pass@1. Thus, it is necessary for LLMs to see more contexts about cross-file dependencies.

\noindent \textbf{The bias of Recall@$k$.}
As stated in Section \ref{sec:bench:metric}, we develop a static analysis-based parser to extract dependencies in generated programs automatically. Because Python is a dynamically typed language, certain dependencies are only determined at runtime and may elude our parser. It may lead to lower Recall@$k$ than actual values.

To gauge the above bias, we randomly select 50 programs generated by gpt-4 and annotate dependencies with them by our parser and two human developers, respectively. Based on the human-annotated and auto-extracted dependencies, we compute two Recall@1 values. The bias of two Recall@1 values is 0.16. Compared to the average variations between LLMs (7.77 in Table \ref{tab:CGen}), 0.16 is slight. Consequently, we believe that Recall@$k$ can effectively rank different LLMs, notwithstanding its slight bias.

\section{Related Work}
\label{sec:related_work}

\noindent \textbf{Large Language Models for Code Generation.}  
The rise of pre-training technology has brought new impetus to the field of code generation, both in academia and industry \cite{alphacode, industry, codegen, incoder}. In this context, more and more LLMs have emerged, achieving significant advancements in code generation, such as Codex \cite{Codex}, ChatGPT \cite{gpt-3.5}, CodeLlama \cite{CodeLLaMa}, DeepSeek Coder \cite{DeepSeek_Coder}, and StarCoder2 \cite{StarCoder-2}. 

To effectively steer LLMs in various code generation scenarios, some works focus on improving the prompt technologies by introducing specific patterns, \eg Structured Chain-of-Thought \cite{SCoT}, Self-planning \cite{Self-planning}, Self-debug \cite{self-debug}, Self-collaboration \cite{Self-collaboration}, and AceCoder \cite{AceCoder}. 

\noindent \textbf{Code Generation Benchmarks.} 
Early code generation benchmarks \cite{CoNaLA,Codex,MBPP,CERT} evaluate code generation on relatively Python functions, such as HumanEval \cite{Codex} and MBPP \cite{MBPP}. APPS \cite{APPS} evaluates code generation on more difficult competition-style problems. ClassEval \cite{ClassEval} evaluates LLMs on class-level code generation and contains 100 human-crafted self-contained Python classes. 
Concode \cite{Concode} and CoderEval \cite{CoderEval} further introduce non-standalone programs.

We release \bench to extend code generation benchmarks. Compared to existing benchmarks, \bench aligns with real-world code repositories (\eg the distributions of code and dependency) and contains more comprehensive annotations (\eg reference dependencies). Besides, \bench is an evolving benchmark and is dynamically updated to address data leakage.

We have also noticed that some benchmarks have recently been proposed for repository-level tasks. CrossCodeEval \cite{CrossCodeEval}, RepoBench \cite{RepoBench}, and RepoEval \cite{RepoCoder} are code completion benchmarks. They lack the necessary annotations (\eg natural language requirements) for code generation. SWE-bench \cite{SWE-bench} focuses on repairing repositories' issues by revising existing programs. In contrast, \bench is collected for code generation and aims to generate new programs based on requirements for a repository. \bench offers comprehensive annotations (\eg natural language requirements, original repositories, reference code, and reference dependencies).

\section{Conclusion and Future Work}
\label{sec:conclusion}

In this paper, we propose a new code generation benchmark named \bench. Collected through a meticulous pipeline, \bench aligns with real-world code repositories in multiple dimensions, \eg code distributions and dependency distributions. Besides, \bench is an evolving benchmark and will be dynamically updated every period (\eg 6 months).
Based on \bench, we propose repository-level code generation and evaluate 10 popular LLMs. The results reveal the strengths and weaknesses of LLMs in real-world repositories. Compared to previous benchmarks, \bench offers a more challenging and realistic evaluation scenario. We hope \bench can facilitate the applications of LLMs in real-world repositories.

In the future, we will continue to update \bench, \eg multilingual samples. Besides, we will explore how to improve the performance of LLMs in repository-level code generation, \eg retrieval-augmented and tool-augmented generation.

\section{Limitations}
\label{sec:limitation}

We believe that \bench itself has four limitations. 
\ding{182} \bench is a monolingual benchmark (\ie requirements in English and code in Python) and ignores other languages. In practice, LLMs require understanding requirements in different natural languages (\eg Chinese, Spanish) and generating programs in various programming languages (\eg Java, C). Thus, we plan to build a multilingual \bench in future work. 
\ding{183} Auto-generated requirements can be improved. As stated in Section \ref{sec:discussion}, auto-generated requirements are comparable to human-written requirements but may lack necessary details. In the future, we will leverage more advanced generation strategies (\eg controlled text generation \cite{control_gen}) to generate requirements.

\ding{184}  As stated in Section \ref{sec:discussion}, Recall@$k$ values in \bench may have slight biases, \ie they may be slightly less than actual values. Because Python is a dynamically typed language, certain dependencies can only be identified at runtime and may elude our parser. 
To gauge the bias introduced by our parser, we manually annotate dependencies within 50 programs generated by gpt-4. Simultaneously, we employ the parser to extract dependencies in the same 50 programs. Based on the human-annotated and auto-extracted dependencies, we compute two Recall@1 values. The bias of two Recall@1 is 0.16. Compared to the average variations between LLMs (7.77 in Table \ref{tab:CGen}), 0.16 is slight. Consequently, the Recall@$k$ can effectively rank different LLMs, notwithstanding its slight bias.

Besides, our evaluation experiments can be further improved in three aspects. 
\ding{185} More LLMs. Due to the limited computing budgets, we mainly evaluate 10 mainstream LLMs. It is worth evaluating LLMs with different sizes and fine-tuned models.
\ding{186} More investigations of contexts. As shown in Table \ref{tab:CGen}, we design two straightforward approaches to extracting code contexts from repositories. In the future, we will introduce cross-file contexts and further explore how to utilize contexts effectively.
\ding{187} Tuning hyper-parameters. It is known that LLMs are sensitive to sampling hyper-parameters and prompts. We ensure all LLMs are evaluated under the same experimental settings. Due to the limited computing budgets, we do not carefully tune hyper-parameters and prompts. Thus, there may be better hyper-parameters and prompts to improve the performance of LLMs further.

\section{Ethics Consideration}
\label{sec:ethics}

\bench is collected from open-source repositories from the real world. We manually check all samples in \bench. We ensure all samples do not contain private information or offensive content. We ensure all programs in \bench are behaving normally and exclude any malicious programs.

\bibliography{anthology,custom}
\bibliographystyle{acl_natbib}

\appendix

\twocolumn

\section{Benchmark Collection Details}
\label{sec:appendix:collection}

\subsection{Requirement Annotation}
\label{sec:appendix:collection:gen_requirement}

\begin{figure}[t]
\centering
\includegraphics[width=\linewidth]{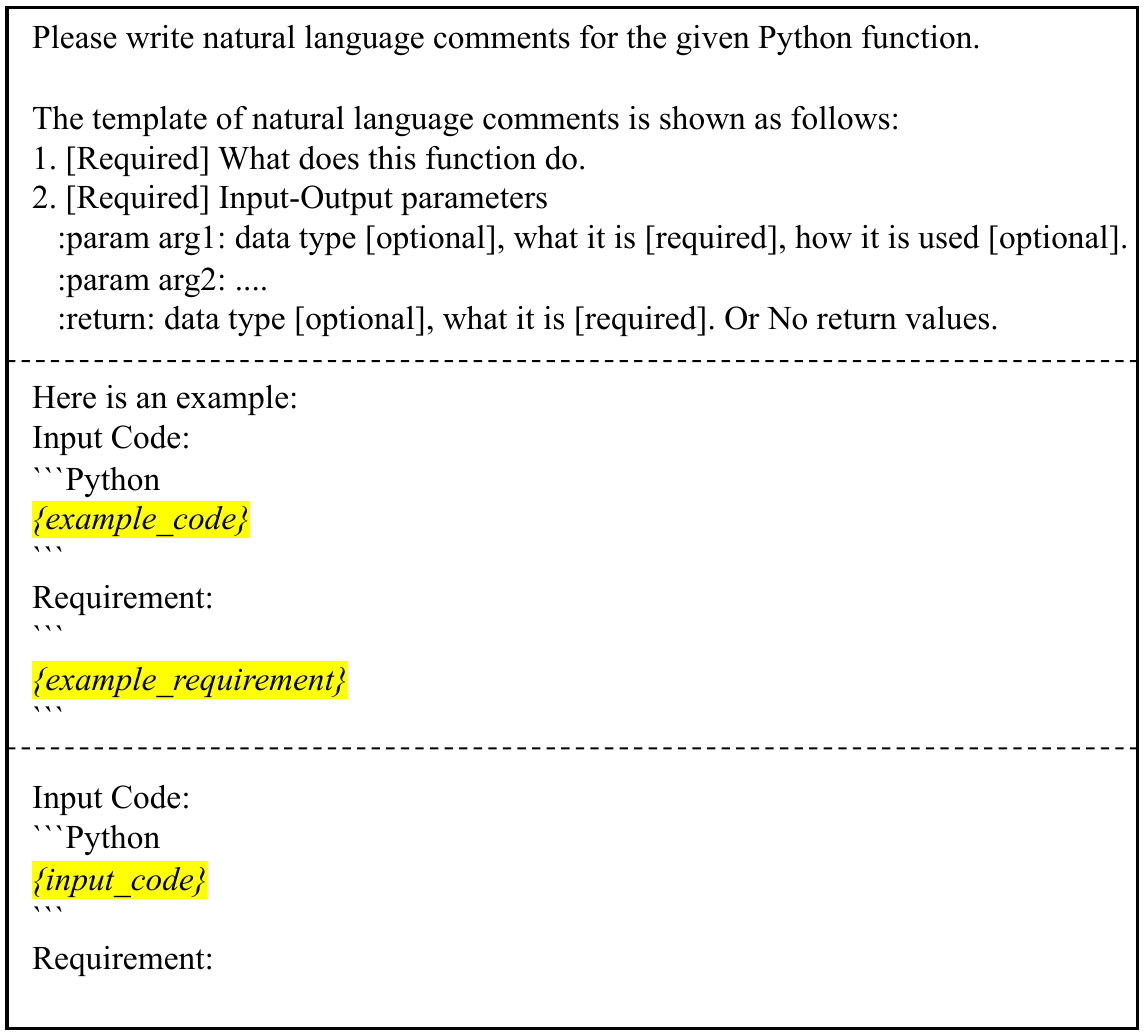}
\caption{The prompt template for generating requirements with gpt-4.}
\label{fig:gen_requirement_prompt}
\end{figure}

Figure \ref{fig:gen_requirement_prompt} shows the prompt template for generating requirements. \texttt{\{example\_code\}}, \texttt{\{example\_requirement\}}, and \texttt{\{input\_code\}} are placeholders. We manually write a function with requirements as the demonstration example. Then, we fill \texttt{\{input\_code\}} with functions and leverage gpt-4 to generate the corresponding requirements. We use the greedy search and generate a requirement for each function.

\subsection{Projects in \bench-2403}
\label{sec:appendix:collection:projects}

\begin{table*}[t]
\centering
\caption{Projects detail on \bench-2403.}
\label{tab:CGen}
\resizebox{0.9\linewidth}{!}{
\begin{tabular}{lcccccc}
\toprule
Repository & Created & Stars & Py Files & Py Lines & Samples & Domin\\
\midrule
Test-Agent & 2023-10-20 & 440 & 85 & 15278 & 1 & Deep Learning \\
skfolio & 2023-12-14 & 813 & 158 & 33852 & 13 & Statistical Learning \\
camp\_zipnerf & 2024-01-19 & 523 & 53 & 18973 & 54 & Image Processing \\
microagents & 2023-12-11 & 674 & 45 & 2918 & 18 & Deep Learning \\
open-iris & 2023-12-09 & 161 & 140 & 13933 & 14 & Image Processing \\
litdata & 2024-02-15 & 114 & 56 & 11713 & 59 & Data Science \\
nlm-ingestor & 2024-01-17 & 643 & 56 & 16674 & 4 & Internet \\
AutoRAG & 2024-01-10 & 259 & 115 & 7735 & 13 & Data Science \\
XAgent & 2023-10-16 & 7054 & 148 & 17623 & 3 & Deep Learning \\
tanuki\_py & 2023-10-16 & 606 & 108 & 10146 & 9 & Text Processing \\
UHGEval & 2023-11-06 & 148 & 34 & 2938 & 3 & Data Science \\
Generalizable-BEV & 2023-10-30 & 136 & 570 & 132407 & 8 & Image Processing \\
EasyVolcap & 2023-12-07 & 442 & 308 & 51723 & 20 & Image Processing \\
UniRef & 2023-12-22 & 208 & 382 & 70042 & 23 & Image Processing \\
contrastors & 2024-01-30 & 346 & 62 & 13774 & 1 & Deep Learning \\
gaussian-splatting-lightning & 2023-10-06 & 168 & 76 & 9935 & 1 & Image Processing \\
scepter & 2023-12-21 & 190 & 244 & 41519 & 1 & Deep Learning \\
microsearch & 2024-02-05 & 336 & 5 & 231 & 2 & Internet  \\
ollama-python & 2023-12-09 & 898 & 13 & 2089 & 12 & Deep Learning \\
Python-Type-Challenges & 2023-10-23 & 343 & 121 & 3208 & 1 & Education \\
stable-fast & 2023-10-17 & 871 & 82 & 11948 & 2 & Deep Learning \\
stable-diffusion-webui-forge & 2024-01-14 & 2537 & 1112 & 210946 & 3 & Deep Learning \\
openlogprobs & 2023-11-22 & 174 & 6 & 524 & 1 & Deep Learning \\
searcharray & 2023-11-03 & 133 & 25 & 4217 & 6 & Data Science \\
deluder & 2023-12-01 & 115 & 34 & 1894 & 3 & Internet \\
\bottomrule
\end{tabular}}
\end{table*}


\section{Experimental Details}
\label{sec:appendix:experiment}

\subsection{Base LLMs}
\label{sec:appendix:experiment:llm}

In this paper, we select 6 popular LLMs as base LLMs and evaluate them on \bench-2403. The details of these LLMs are described as follows.
\begin{itemize}
    \item \textbf{gpt-4} \cite{gpt-4}, released by OpenAI on March 14, 2023, marks another milestone in the field of natural language processing. gpt-4 demonstrates superior performance compared to previous gpt models \cite{DBLP:journals/corr/abs-2303-12712}. In our experiments, we use the version - gpt-4-1106. Its training data up to April 2023. It continues the auto-regressive prediction of the next token training objective inherited from the GPT series models. It incorporates reinforcement learning with human feedback (RLHF) and red-teaming \cite{red-teaming} techniques. However, the pre-training data scope and scale, model size, and parameters remain closed-source at present.
    
    \item \textbf{gpt-3.5-turbo} \cite{gpt-3.5} is an improved gpt-3 model enhanced by a three-stage reinforcement learning with human feedback (RLHF) algorithm. Apart from improving instruction-following capabilities, the RLHF algorithm proves highly effective in mitigating the generation of harmful or toxic content, which is crucial for the practical deployment of LLMs in security-sensitive contexts. we utilized the released versions of gpt-3.5, namely gpt-3.5-turbo-1106, with training data up to September 2021. However, similar to gpt-4, the training details, training data, and model weights are currently closed-source.  

    \item \textbf{CodeLLaMa} \cite{CodeLLaMa}, based on the LLama2 architecture by Meta-AI\footnote{\url{https://ai.meta.com/}}, was fine-tuned and open-sourced by the company on August 25, 2023, with versions of 7B, 13B, and 34B. A 70B version was released on January 30, 2024 \cite{CodeLLaMa}. CodeLLama is primarily trained on nearly deduplicated publicly available code datasets. The first three models were trained on 500 billion tokenized code, while the latest 70B model was trained on 1T tokens. Similar to the LLaMa series, CodeLLaMa also follows a decoder-only architecture. We evaluated CodeLLaMa-Python-\{7B, 13B\} upon our \bench.
       
    \item \textbf{DeepSeek Coder} \cite{DeepSeek_Coder} is a large language model for programming tasks released by DeepSeek-AI\footnote{\url{https://www.deepseek.com/}} in November 2, 2023. DeepSeek Coder consists of a series of code language models, each trained from scratch on 2T tokens, containing 87\% code and 13\% natural language. DeepSeek Coder provides code models with 1.3B, 6.7B and 33B parameter sizes. In terms of model architecture, each model integrates a decoder-only Transformer, incorporating Rotary Position Embedding and FlashAttention v2. We evaluated DeepSeek Coder-\{6.7B, 33B\} on our \bench.

    \item  \textbf{StarCoder 2} \cite{StarCoder-2} was released by BigCode\footnote{\url{https://www.bigcode-project.org/}} on December 8, 2023 with 3 different parameters, 3B, 7B and 15B. StarCoder2 is trained on The Stack v2, a new large-scale, high-quality code dataset. All models were trained using Grouped Query Attention, a contextual window of 16,384 tokens with a sliding window attention of 4,096 tokens, using the Fill-in-the-Middle objective. Following DeepseekCoder \cite{DeepSeek_Coder} and Code LLaMA \cite{CodeLLaMa}, StarCoder2 use Rotary Positional Encodings. We evaluated StarCoder2-\{7B, 15B\} on our \bench, which was trained on over 3.5 trillion tokens in 17 programming languages from Stack v2.

    \item  \textbf{Gemma} \cite{gemma} is a lightweight open model family built from the research and technology used to create Gemini models. Gemma models demonstrate strong performance on academic benchmarks in language understanding, reasoning, and security. Gemma releases models in two sizes (2 billion and 7 billion parameters) and provides pre-training and fine-tuning checkpoints. Gemma is released by Google DeepMind\footnote{\url{https://deepmind.google/}} on February 21, 2024, based on the transformer decoder. Models are trained on a context length of 8192 tokens. We evaluated Gemma-7b in our \bench performance.

    \item  \textbf{Qwen} \cite{Qwen} is a comprehensive language model series containing different models with different parameters, including basic pre-trained language models and chat models fine-tuned through human alignment technology. The Qwen series is open-sourced, including Qwen-1.8B, Qwen-7B, Qwen-14B, and Qwen-72B, and the corresponding Qwen-Chat series, which the Alibaba Group released on August 3, 2023. Qwen has conducted stable pre-training on multi-lingual data of up to 3 trillion tokens, covering fields, languages (focusing on Chinese and English), etc. QWEN is designed using a modified version of the Transformer architecture, specifically adopting the open-source approach of LLaMA \cite{LLaMa}. We evaluated the performance of Qwen1.5-7b on our \bench.
    
\end{itemize}

\subsection{Prompt Templates}
\label{sec:appendix:experiment:prompt}

\begin{figure}[t]
\centering
\includegraphics[width=\linewidth]{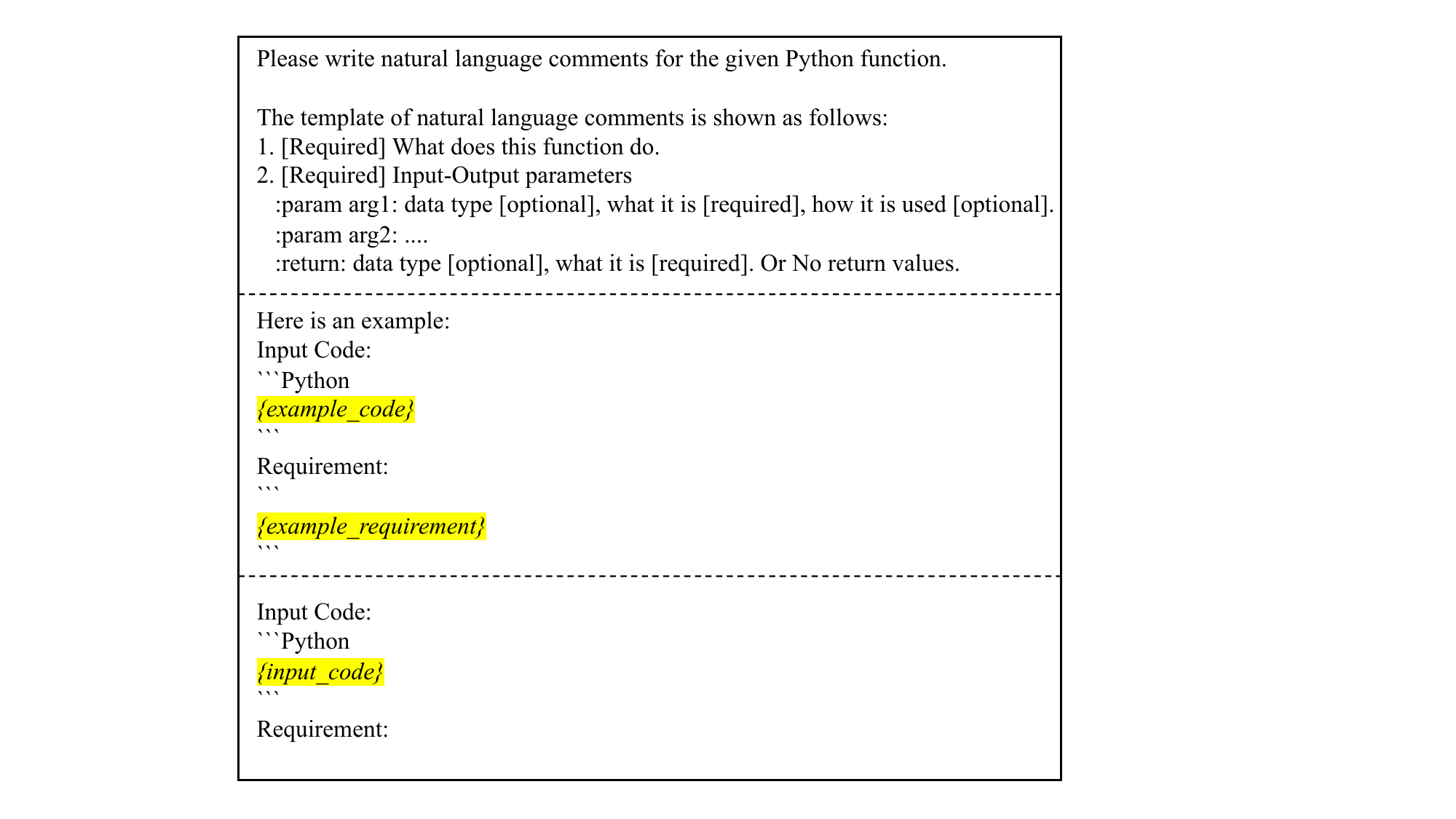}
\caption{The prompt template in the without context setting.}
\label{fig:nl_prompt}
\end{figure}

\begin{figure}[t]
\centering
\includegraphics[width=\linewidth]{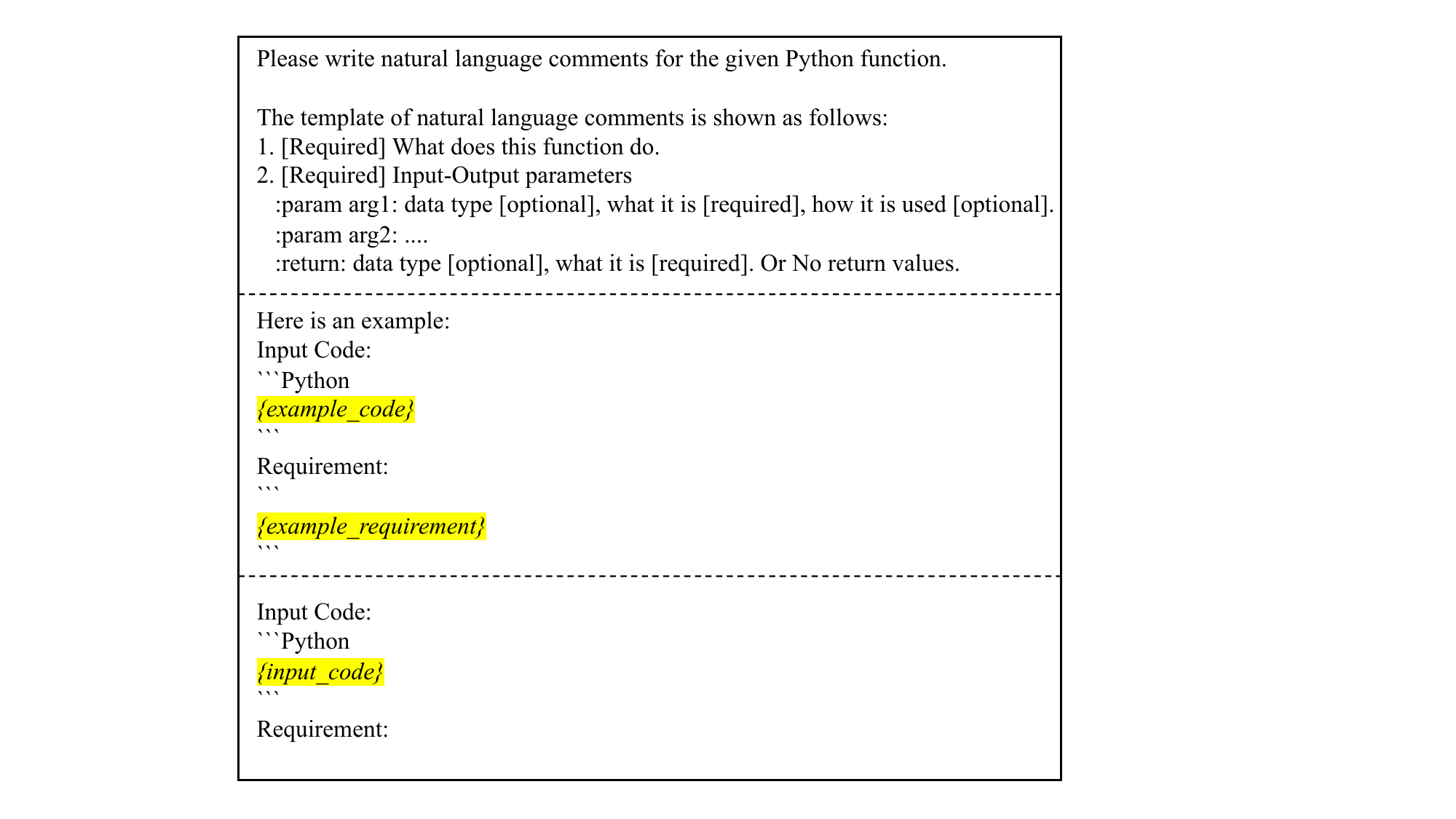}
\caption{The prompt template in the local file (completion) setting.}
\label{fig:local_completion_prompt}
\end{figure}

\begin{figure}[t]
\centering
\includegraphics[width=\linewidth]{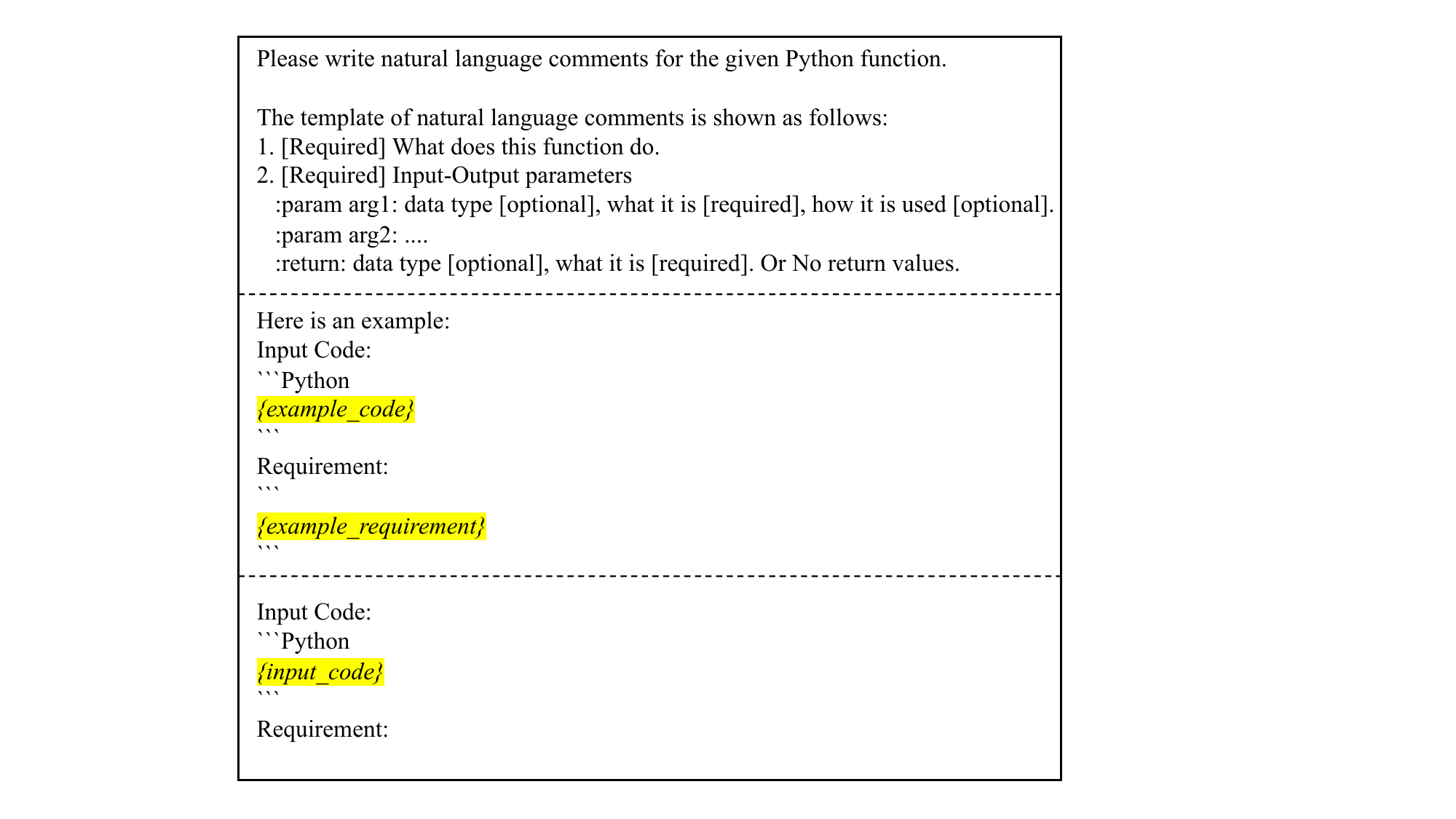}
\caption{The prompt template in the local file (infilling) setting.}
\label{fig:local_infilling_prompt}
\end{figure}

\begin{figure}[t]
\centering
\includegraphics[width=\linewidth]{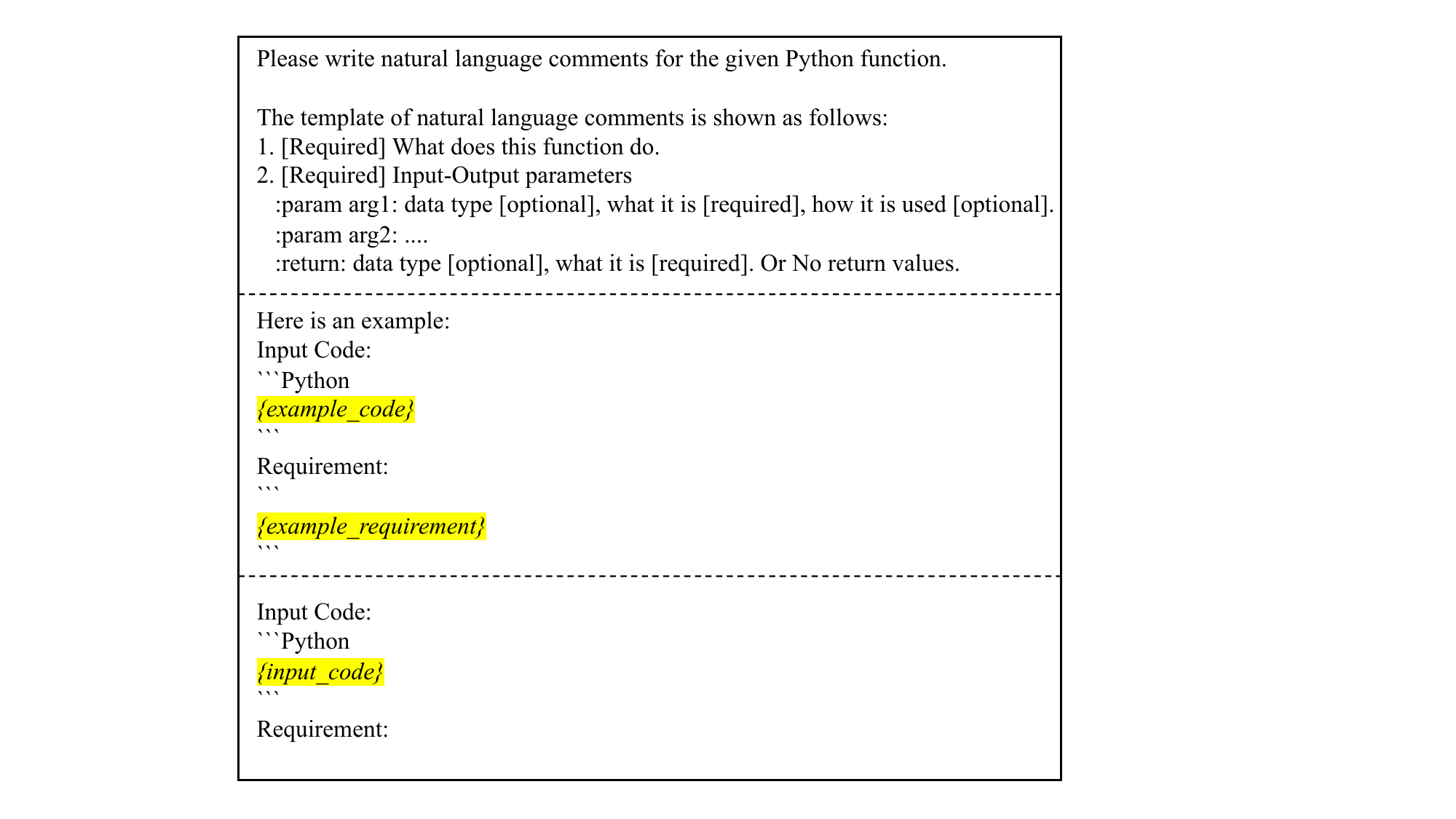}
\caption{The prompt template in the similar function setting.}
\label{fig:similar_func_prompt}
\end{figure}

The prompt templates used for instruction-tuning models (\ie gpt-4 and gpt-3.5) are shown in Figure \ref{fig:nl_prompt}, \ref{fig:local_completion_prompt}, \ref{fig:local_infilling_prompt}, and \ref{fig:similar_func_prompt}. \texttt{\{function\_name\}}, \texttt{\{contexts\_above\}}, \texttt{\{contexts\_below\}}, \texttt{\{signature\}}, and \texttt{\{requirement\}} are placeholders.

For other standard language models, the prompt templates are shown as follows: \ding{172} Without context: [\texttt{signature; requirement}]; \ding{173} Local file (completion): [\texttt{context\_above; signature; requirement}]; \ding{174} Local file (infilling): [\texttt{prefix\_id; context\_above; signature; requirement; suffix\_id; context\_below; middle\_id}].
Where \texttt{[;]} denotes the concatenation operation of strings. \texttt{\{prefix\_id\}}, \texttt{\{suffix\_id\}}, \texttt{\{middle\_id\}} are special tokens used in code infilling. For different LLMs, we reuse their official special tokens to make prompts. 

\end{document}